\pdfoutput=1

\documentclass[11pt]{article}

\usepackage{acl}
\usepackage{amsmath,amsfonts,bm}

\usepackage{amsmath,amsfonts,bm}
\usepackage{amsthm}

\def\figref#1{figure~\ref{#1}}

\def\tblref#1{table~\ref{#1}}

\def\eqref#1{equation~\ref{#1}}
\def\Eqref#1{Equation~\ref{#1}}

\def\1{\bm{1}}

\def\ve{{\bm{e}}}

\def\vh{{\bm{h}}}

\def\vp{{\bm{p}}}

\def\vv{{\bm{v}}}
\def\vw{{\bm{w}}}
\def\vx{{\bm{x}}}
\def\vy{{\bm{y}}}

\DeclareMathAlphabet{\mathsfit}{\encodingdefault}{\sfdefault}{m}{sl}
\SetMathAlphabet{\mathsfit}{bold}{\encodingdefault}{\sfdefault}{bx}{n}

\def\gL{{\mathcal{L}}}

\def\sR{{\mathbb{R}}}

\newcommand{\sigmoid}{\sigma}

\usepackage{bbm}
\usepackage{subcaption}
\usepackage{xcolor}
\usepackage{makecell} 
\usepackage{adjustbox}

\usepackage{bbold}
\usepackage[most]{tcolorbox}
\usepackage{comment}
\usepackage{multirow}
\usepackage{multicol}
\usepackage{booktabs}

\usepackage{bbm}
\usepackage{times}
\usepackage{latexsym}

\usepackage{xcolor}
\definecolor{yaushian}{RGB}{243, 101, 66}
\definecolor{ruohong}{RGB}{101, 66, 243}
\definecolor{yiming}{RGB}{255, 0, 0}

\usepackage[T1]{fontenc}

\usepackage[utf8]{inputenc}

\usepackage{microtype}

\newcommand{\ourmodel}{\textsc{GlocalXML }}
\newcommand{\ourmodelshort}{\textsc{GlocalXML}}

\title{Exploiting Local and Global Features in Transformer-based \\ Extreme Multi-label Text Classification}
\makeatletter
\newcommand{\printfnsymbol}[1]{%
  \textsuperscript{\@fnsymbol{#1}}%
}
\makeatother

\author{Ruohong Zhang\thanks{\quad The first two authors contribute equally.} \and Yau-Shian Wang\footnotemark[1] \\
  \texttt{ruohongz,yaushiaw@andrew.cmu.edu} \\\And
  Yiming Yang \\
  \texttt{yiming@cs.cmu.edu} \\ \AND
  Tom Vu \\
  \texttt{tom.m.vu@gmail.com} \\\And 
  Likun Lei \\
  \texttt{llei@flexport.com}
  }

\begin{document}
\maketitle
\begin{abstract}
Extreme multi-label text classification (XMTC) is the task of tagging each document with the relevant labels from a very large space of predefined categories. Recently, large pre-trained Transformer models have made significant performance improvements in XMTC, which typically use the embedding of the special CLS token to represent the entire document semantics as a global feature vector, and match it
against candidate labels. However, 
we argue that such a global feature vector may not be sufficient
to represent different granularity levels of semantics in the document, and that complementing it with the local word-level features 
could bring additional gains. Based on this insight, we propose an 
approach that combines both the local and global features produced by Transformer models to improve the prediction power of the classifier.
Our experiments show that the proposed model either outperforms or is comparable to the state-of-the-art methods on benchmark datasets.
\end{abstract}
\section{Introduction}
Extreme multi-label text classification (XMTC) is the task of tagging each document with relevant labels 
where the target space may contains up to thousands of category labels.  Those labels typically form a semantic hierarchy, where the higher-level labels correspond to more abstract or general concepts, while the lower-level labels specify fine-grained distinctions.
XMTC has many real-world applications, such as assigning subject topics to news or Wikipedia articles, tagging keywords for online shopping items, classifying industrial products for tax purposes, and so on. 

A central problem in XMTC is to learn a good representation of each input document that well-captures the semantic information for label predictions. 
Traditional classifiers typically use the bag-of-words (BoW) representation, which is a vector of features (words) with
TF (term frequency within a document) and IDF (the inverse document frequency in a document collection) weights. 
As word location, ordering and semantic dependencies in context are ignored, the BoW representation is sub-optimal for capturing the contextualized semantic information of the input document. 
This limitation has been addressed by the recent neural network approaches with the ability of  
learning of contextual embeddings of documents, 
especially with large pre-trained Transformer-based models such as BERT~\cite{devlin2018bert}, Roberta \cite{liu2017deep} and XLNet~\cite{yang2019xlnet}. 
Successful examples of such neural XMTC solvers include X-Transformer~\cite{chang2020taming}, APLC-XLNet~\cite{ye2020pretrained} and LightXML~\cite{jiang2021lightxml}, which achieved state-of-the-art (SOTA) performance on several benchmark datasets.

In those Transformer-based models, the embedding of token [CLS] at a each layer of the neural network summarizes the content of the input document into a single vector. 
As those vectors reflect the semantic of document as a whole, we call them the \textit{global features} in our paper.
Usually the [CLS] embeddings at the last~\cite{chang2020taming, ye2020pretrained} or last few layers~\cite{jiang2021lightxml} are used for the label prediction because the global semantics of those are enriched from
 multi-layers of self-attention. 
 While using those global features for label predictions is a natural choice, we argue that it may not be sufficient for fully exploiting the advantages of Transformer models. Specifically, 
the embeddings of all word tokens can directly participate in label prediction. 
Following this intuition, we study how to  
leverage the word embeddings (especially from a lower layer of) Transformer in addition to the global features in XMTC. As the word embeddings reflect finer details of a document compared to the global summarization, we call them the \textit{local features}.


We argue that the different labels in XMTC can reflect the semantic contents of a document at various granularity levels. As an intuitive example, the Amazon product "Falling in Love Is Wonderful" is a collection of Broadway love duets. The category "music" can be inferred from the global feature summarizing the content of the product. On the other hand, 
the finer-grained categories such as "vocalist" and "soundtracks" are easier to be predicted directly from the keywords "singer" and "recording" in the text description. 
The important signals carried by "singer" and "recording" could be overshadowed in the global summary of the full document if we only focus on its global features, i.e., the [CLS] embeddings at the last or last few layers.
This leads to our key idea of directly leveraging the local word level features for label prediction, especially those from the lower layers of Transformer models. Specifically, we provide a method that lets each label attentively select the keywords in the document text with the label-word attention. This method puts more emphasis on the matching 
between each label and document words, complementing the global features with the details in context.

To build a robust and efficient classification system, we propose an integrating framework that combines both the local and the global features in pre-trained Transformer models, namely \ourmodelshort. Our experiments demonstrate the effectiveness of our proposed method which either outperforms or is comparable to the SOTA model the benchmark XMTC datasets. We also conduct ablation studies to verify the effectiveness of our model in utilizing both features.

\section{Proposed Method} \label{sec:method}
\subsection{Preliminaries}
Let $\vx = \{ x_1, x_2, \ldots, x_T \}$ be the input document with length $T$, and the set of associated ground truth labels is $\vy \in \sR^L$ with $y_l \in \{0, 1\}$, where $L$ is the label size. A classier calculates a probability
$p_l$ of the label being true.
The binary cross entropy (BCE) loss between $\vp= \{ p_1, p_2, \ldots, p_L \}$ and $\vy$ is calculated as:
\begingroup
\small
\begin{equation*}
    \gL_{\text{BCE}}(\vp, \vy) = -\frac{1}{L}\sum_{l \in L} \bigg [ y_l\log p_l + (1-y_l)\log (1-p_l) \bigg ]\,.
\end{equation*}
\endgroup

The document $x$ is usually prepended with a special [CLS] token before input to the Transformer model. For a Transformer model with $N$ layers, the hidden representations from the n-th layer is denoted as:  
\begin{equation}
    \phi_{\text{transformer}}^{(n)}(\vx)=\{\vh^{(n)}_{cls}, \vh^{(n)}_1, \vh^{(n)}_2,..., \vh^{(n)}_T\}\,.
\end{equation}
We will introduce our classification system with global and local features respectively.

\subsection{Classification with Global Features} 
Global features denote the [CLS] embedding summarizing he high-level and abstract representations of document content. 
We use the [CLS] embedding from the last layer $N$ because it contains the richest information after multiple layers of self-attention. 
Formally, we use $\vh^{(N)}_{cls}$ (or optionally passed to a linear pooler) as the global feature. The probability of predicting a label $l$ is calculated by:
\begin{equation}
    p_l^{\text{global}} = \sigmoid(\langle \vh^{(N)}_{cls}, \ve_l^{\text{global}} \rangle)\,, \label{eq:global}
\end{equation}
where $\ve_l^{\text{global}}$ is the label embedding for global features and $\langle \cdot, \cdot \rangle$ is the dot product.
The classifier with global feature directly maps the document representation against the label representation.


\subsection{Classification with Local Features} 
The local features denote all the token embeddings at a certain layer of Transformer, which preserve the diverse and fine-grained token information peculiar to the label of interest. As the first layer token embeddings from a Transformer model mostly pertain to the token surface-level meaning (while being contextualized), we select them as the local features. 

Similar to label-word attention~\cite{you2018attentionxml}, our model is designed to let each label attentively select the key tokens from the document.
Specifically, we treat labels as queries to retrieve the salient tokens in the documents ($\vh^{(1)}_{cls}$ is written as $\vh^{(1)}_{0}$):
\begin{align}
    & \operatorname{\psi_K}(\phi_{\text{transformer}}^{(1)}(\vx)) = \{\vw_1^k, \vw_2^k, \ldots, \vw_T^k \}\,, \\
    & \alpha_{ij} =  
    \frac{ \exp( \langle \vw^k_i , \ve_j^{local} \rangle / \tau) }
    {\sum_{t=0}^T \exp (\langle\vw^k_t, \ve_j^{local} \rangle / \tau)}\, , \label{eq:local}
\end{align}
where $\psi_K$ is a linear function, 
$\ve_j^{local}$ is the label embedding for local features and $\tau$ is the temperature.
$\tau$ controls the smoothness of the attention distribution over the words. With a smaller $\tau < 1$, the attention is peaked on the most salient key tokens.


\Eqref{eq:global} and \ref{eq:local} highlights the difference between the usage of global and local features: for the global features, relevance scores are computed between the label embeddings and the document embedding, while for the local features, relevance scores are directly computed between the label embeddings and the token embeddings for key token selection. 

The retrieved key tokens are aggregated according to the relevance score $\alpha_{ij}$:
\begin{align}
    & \operatorname{\psi_V}(\phi_{\text{transformer}}^{(1)}(\vx)) = \{\vw_1^v, \vw_2^v, \ldots, \vw_T^v \} \,, \\
    &\vv_j = \sum_{i=0}^T \alpha_{ij}\vw^v_i\,, \quad 
    p_j^{\text{local}} = \sigmoid(\phi_{MLP}(\vv_j))\,,
\end{align}
where $\psi_V$ is a linear function and $\phi_{MLP}(\vv_j)$
is the multi-layer perceptron that summarizes the aggregated token embedding into a real-valued score.





\subsection{Training and Inference}
\subsubsection{Inference}
Our framework integrates the classification with local and global features, and the final prediction for a given input text and label $l$ is:
\begin{equation}
    p_l^\text{final} = \frac{1}{2}(p_l^{\text{local}} + p_l^{\text{global}})\,.
\end{equation}

\subsubsection{Training Objective}
Instead of optimizing $p_l^\text{final}$ directly, we optimize $\vp^{\text{global}}$ and $\vp^{\text{local}}$ by independent losses:
\begin{equation}
    \gL_{\text{total}} = \gL_{\text{BCE}}(\vp^{\text{global}}, \vy) + \gL_{\text{BCE}}(\vp^{\text{local}}, \vy)\,.
\end{equation}
Optimizing the two classifiers separately encourages each module to focus on its own specialities. As the backbone of Transformer is shared, the unified framework allows us to build a fast and robust model with wider applicability.

\begin{table}[th!]
    \small
    \caption{Corpus Statistics: $N_{\text{train}}$ and $N_{\text{test}}$ are the number of training and testing instances respectively; $\bar{L}_d$ is the average number of labels per document, and $L$ is the number of unique labels. \label{tab:dataset}}  
    \begin{adjustbox}{width=\columnwidth,center}
    \begin{tabular}{l|ccccc}
    \toprule
    Dataset & $N_{train}$ & $N_{test}$  & $\bar{L}_d$ & $L$  \\
    \midrule
    EURLex-4K & 15,539 &  3,809  & 5.30 & 3,956  \\
    Wiki10-31K & 14,146 & 6,616  & 18.64 &  30,938  \\
    AmazonCat-13K & 1,186,239 & 306,782  & 5.04 & 13,330  \\
    \bottomrule
    \end{tabular}
    \end{adjustbox}
\end{table}

\begin{table*}[ht]
    \centering
     \caption{The prediction results of representative classification systems evaluated in the micro-avg P@k metric. The bold phase and underscore highlight the best and second best model performance.  \label{tab:main}}
    \begin{adjustbox}{width=\linewidth,center}
    \begin{tabular}{cc|ccc|ccc|ccc}
    \toprule
    \multicolumn{2}{c}{} & \multicolumn{3}{c}{EURLex-4K} &  \multicolumn{3}{c}{Wiki10-31K} & \multicolumn{3}{c}{AmazonCat-13K} \\
    \midrule
    & Methods & P@1 & P@3 & P@5 & P@1 & P@3 & P@5 & P@1 & P@3 & P@5\\
    \midrule
    \multirow{5}{*}{Statistical models} & DisMEC  &  83.21 & 70.39 & 58.73 & 84.13 & 74.72 & 65.94 & 93.81 & 79.08 & 64.06 \\
    & PfastreXML & 73.14 & 60.16 & 50.54 &  83.57 & 68.61 & 59.10 & 91.75 & 77.97 & 63.68 \\
    & eXtremeText & 79.17 & 66.80 & 56.09 & 83.66 & 73.28 & 64.51 & 92.50 & 78.12 & 63.51 \\
    & Parabel & 82.12 & 68.91 & 57.89 & 84.19 & 72.46 & 63.37 & 93.02 & 79.14 & 64.51 \\
    & Bonsai & 82.30 & 69.55 & 58.35 &  84.52 & 73.76 & 64.69 &  92.98 & 79.13 & 64.46 \\
    \midrule
   \multirow{5}{*}{Neural models} &XML-CNN & 75.32 & 60.14 & 49.21 & 81.41 & 66.23 & 56.11 & 93.26 & 77.06 & 61.40 \\
   & AttentionXML & 87.12 & 73.99 & 61.92 & 87.47 & 78.48 & 69.37 & 95.92 & 82.41 & 67.31 \\
   & X-Transformer & 87.22 & 75.12 & 62.90 & 88.51 & 78.71 & 69.62 & \underline{96.70} & 83.85 & 68.58 \\
   & APLC-XLNet & \underline {87.72} & 74.56 & 62.28 & 89.44 & 78.93 & 69.73 & 94.56 & 79.82 & 64.60\\
   & LightXML & 87.63 & \underline {75.89} & \underline {63.36} & \underline{89.45} & \underline{78.96} & \underline{69.85} & \bf 96.77 & \bf 84.02 & \underline{68.70}\\
   \midrule
   Our model & \ourmodel & \bf 90.32 & \bf 78.90 & \bf 66.20 & \bf 90.11 & \bf 80.95 & \bf 71.97 & 96.60 & \underline{83.97} & \bf 68.78 \\
    \midrule
  \end{tabular}
  \end{adjustbox}
\end{table*}

\section{Experiments}
\subsection{Datasets}
We conduct our experiments on $3$ benchmark datasets: EURLex-4K~\citep{10.1007/978-3-540-87481-2_4}, Wiki10-31K~\citep{zubiaga2012enhancing} and AmazonCat-13K~\citep{10.1145/2507157.2507163}.
The statistics of the datasets are shown in Table~\ref{tab:dataset}. An unstemmed version of EURLex-4K is obtained from the APLC-XLNet github\footnote{\url{https://github.com/huiyegit/APLC_XLNet.git}} and the other two are from the Extreme classification Repository\footnote{\url{http://manikvarma.org/downloads/XC/XMLRepository.html}}. The EURLex-4K is in the European legal domain, the Wiki10-31K is in general domain and the AmazonCat-13K is about product descriptions.

As adapting our proposed method on large XMTC datasets requires extra tree-based techniques such as the two-stage training in X-Transformer \cite{chang2020taming},
we leave that to the future work.

\subsection{Experimental Settings}
\subsubsection{Implementation Details}
As our method can be applied to any Transformer architecture, we use three different pre-trained Transformers base models for each dataset: BERT \cite{devlin2018bert}, Roberta \cite{liu2019roberta} and XLNet \cite{yang2019xlnet}. We report the ensemble score following the experimental settings in previous works \cite{jiang2021lightxml, chang2020taming}.

Following APLC-XLNet~\cite{ye2020pretrained}, we use a smaller learning rate for the pre-trained Transformer backbone module because it may require less tuning.
For the classifier with global features, our implementation uses different learning rate for the Transformer backbone, the pooler (optional) and the classifier, which is $1e-5, 1e-4, 1e-3$ for the Wiki10-31K and $5e-5, 2e-4, 2e-3$ for the other two datasets. For the classifier with local features, we use learning rates of $2e-4, 2e-3$ for the attention module and MLP respectively. We use the fp16 training to reduce the memory usage and increase training speed. We used sequence length of $512$ for BERT and Roberta on EURLex-4K and Wiki10-31K, and $256$ for AmazonCat-13K and the XLNet model. 

\subsubsection{Baselines} We compare our model with the statistical and neural baselines. The statistical models include one-vs-all DisMEC \cite{babbar2017dismec}, PfastreXML \cite{jain2016extreme}; tree-based Parabel \cite{prabhu2018parabel}, eXtremeText \cite{wydmuch2018no}. The deep learning approaches include XML-CNN \cite{liu2017deep}, AttentionXML \cite{you2018attentionxml}; SOTA pre-trained Transformer models X-Transformer \cite{chang2020taming}, APLC-XLNet \cite{ye2020pretrained} and  LightXML \cite{jiang2021lightxml}. 

\subsubsection{Evaluation Metrics}
Following previous work \cite{jiang2021lightxml, chang2020taming}, we evaluate our method with the micro-averaged P@k, which is the most widely-used evaluation metric for XMTC:
\begin{equation}
    P@k = \frac{1}{k} \sum_{i=1}^{k} \mathbbm{1}_{\vy_i^+}(p_i)
\end{equation}
where  $p_i$ is the $i$-th label in a ranked list $\vp$ and $\mathbbm{1}_{\vy_i^+}$ is the indicator function.

\subsection{Main Result}
The performance of model evaluated on the micro-averaged P@k metric is reported in \tblref{tab:main}.
Our model is compared against the statistical and neural models, with the best performance in bold phase and the second best underlined. 

The most competitive baselines are the pre-trained transformer-based models. Our model outperforms those SOTA models on EURLex-4K and Wiki10-31K by a large margin with more than $2\%$ improvement on P@5, and achieves competitive performance on the AmazonCat-13K dataset.
We attribute the performance gains to the usage of local feature in Transformers.
Labels with more specific categorization may directly benefit from attending to token embeddings in the Transformer, when the [CLS] embedding fails to capture the fine-grained details. As for the reason why the gain in AmazonCat is only marginal, that is probably because there are much more training instances in the dataset, so the other neural models have a better chance to encode more useful information into the global embedding, making the advantage of injecting the local information in \ourmodel less obvious.

\begin{table}[ht!]
    \centering
    \caption{Ablation test results for \ourmodel with a single model initialized with Roberta. The performance for the \ourmodelshort, local and glocal classifiers is reported separately. \label{tab:ablation}} 
    \begin{adjustbox}{width=\columnwidth,center}
    \begin{tabular}{lcccc}
    \toprule
    Dataset & & Global  & Local & \ourmodelshort \\
    \midrule
    \multirow{3}{*}{EURLex-4K} 
     & P@1 & 87.27 & 85.98 & 88.93  \\
     & P@3 & 75.09 & 73.36 & 76.90 \\
     & P@5 & 62.97 & 60.76 & 64.22 \\
     \hline
    \multirow{3}{*}{Wiki10-31K} 
     & P@1 & 87.62 & 85.31 & 89.60  \\
     & P@3 & 77.00 & 75.49 & 80.17\\
     & P@5 & 68.25 & 66.92 & 70.99 \\
     \hline
    \multirow{3}{*}{AmazonCat-13K} 
     & P@1 & 96.27 & 95.08 & 96.27 \\
     & P@3 & 83.25 & 81.39 & 83.40 \\
     & P@5 & 68.09 & 66.26 & 68.25 \\
    \bottomrule
    \end{tabular}
    \end{adjustbox}
\end{table}

\begin{figure}[th!]
  \centering
  \begin{subfigure}[b]{0.8\linewidth}
         \centering
         \includegraphics[width=\linewidth]{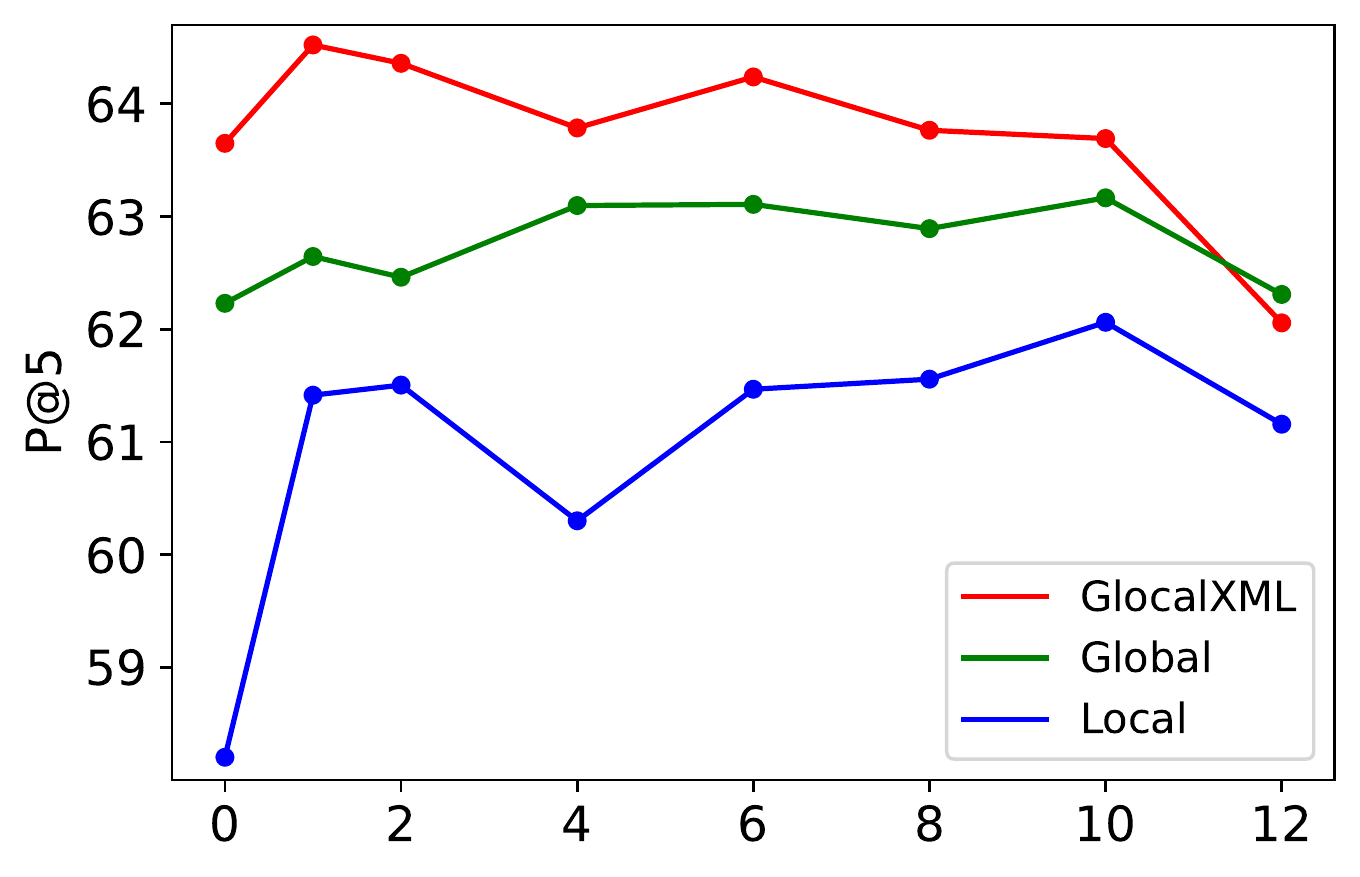}
         \caption{EURLex-4K}
    \end{subfigure}
    \hfill
     \begin{subfigure}[b]{0.8\linewidth}
         \centering
         \includegraphics[width=\linewidth]{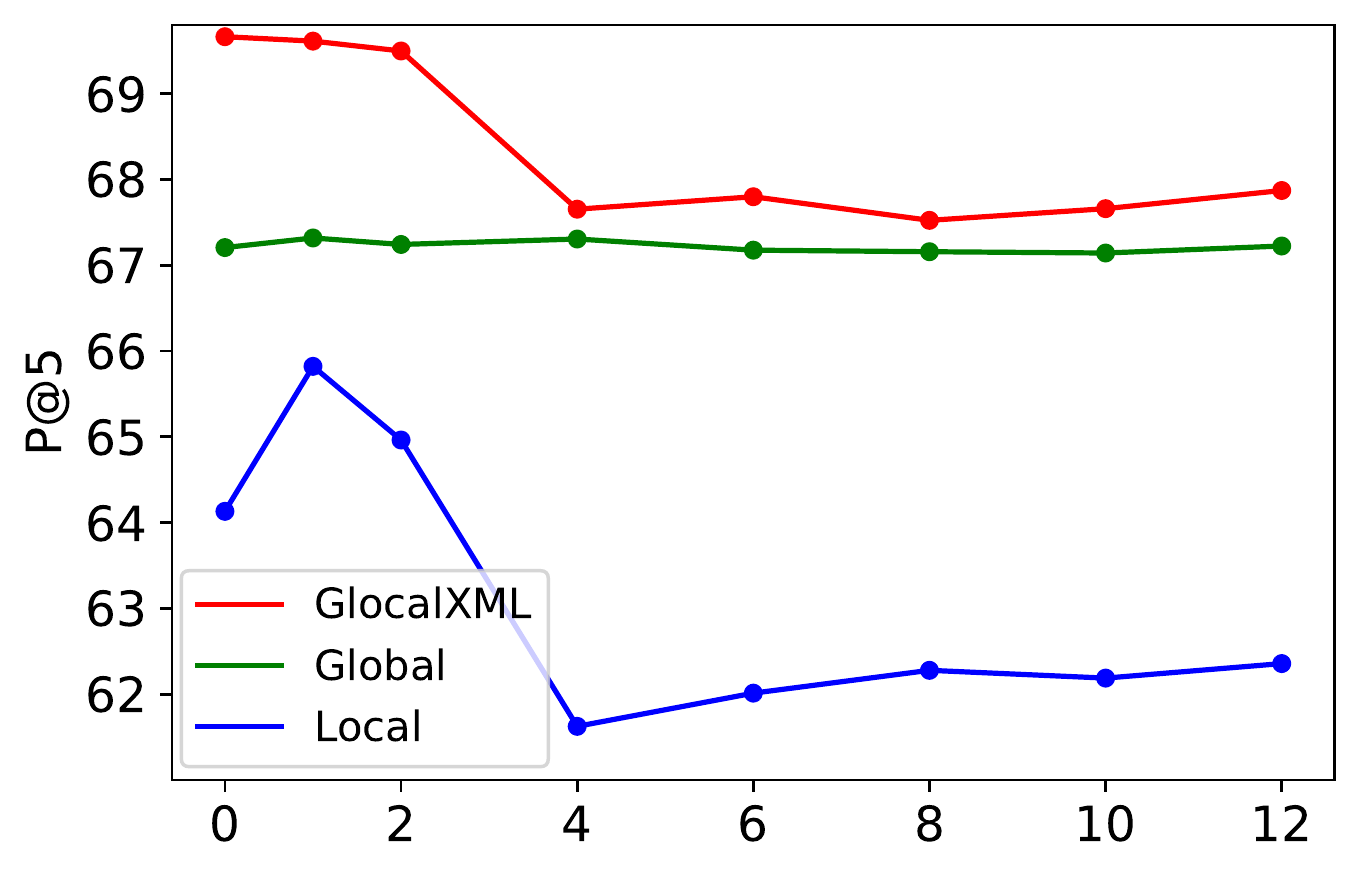}
         \caption{Wiki10-31K}
    \end{subfigure}
  \caption{Ablation test on the effectiveness of combining the global feature ([CLS] embedding at the final layer) with different layers of local feature. The horizontal axis is the local layer number, the vertical axis is the P@5 performance. Layer $0$ corresponds to the original token embeddings.}
  \label{fig:ablation}
\end{figure}
\section{Analysis on Local \& Global Features}
\subsection{Single Model Performance}
The performance of a single Roberta model is reported in \tblref{tab:ablation}. The classifier with local feature inevitably underperforms that with the global feature, probably because the local classifier only shares a shallow Transformer backbone which is less expressive. Despite that, when the local and global features are combined, \ourmodel achieves the best performance, which could come from the complementary effect of local and global feature (analysed below).

\subsection{Local Features on Different Layers}
In \figref{fig:ablation}, we show the performance of combining the global feature with different layers of local features. For efficiency considerations, we reduce the sequence length to $256$ for the experiments. The global feature uses the [CLS] embedding at the final layer of Roberta, while the local features are the token embeddings from Transformer layers $0-12$. The layer $0$ corresponds to the original token embedding without being passed to any Transformer block. 

We observe that the performance of classification with the global feature is relatively stable which outperforms that with the local features. Our \ourmodel model with a combination of the two features achieves the best performance. For the Wiki10-31K dataset, \ourmodel achieves better performance when the features comes from layers $<3$, even with the original token embedding at layer $0$. The reason is that since this dataset has a large label space and each document has an average of $19$ labels, it is more difficult for the [CLS] embedding to summarize the text with distinctive word-level features peculiar to the labels. Therefore, the local classifier which allows labels to directly query for the keywords in the document could pick up the missing information. Combining the two leads to better results.

On both datasets, we observe that the performance of \ourmodel becomes worse when combined with the local features from higher layers, even if the local features from higher layers tend to perform better in EURLex-4K. 
The performance of \ourmodel is peaked when the local feature is at (near) layer $1$.

\begin{figure}[ht]
  \centering
  \includegraphics[width=0.75\linewidth]{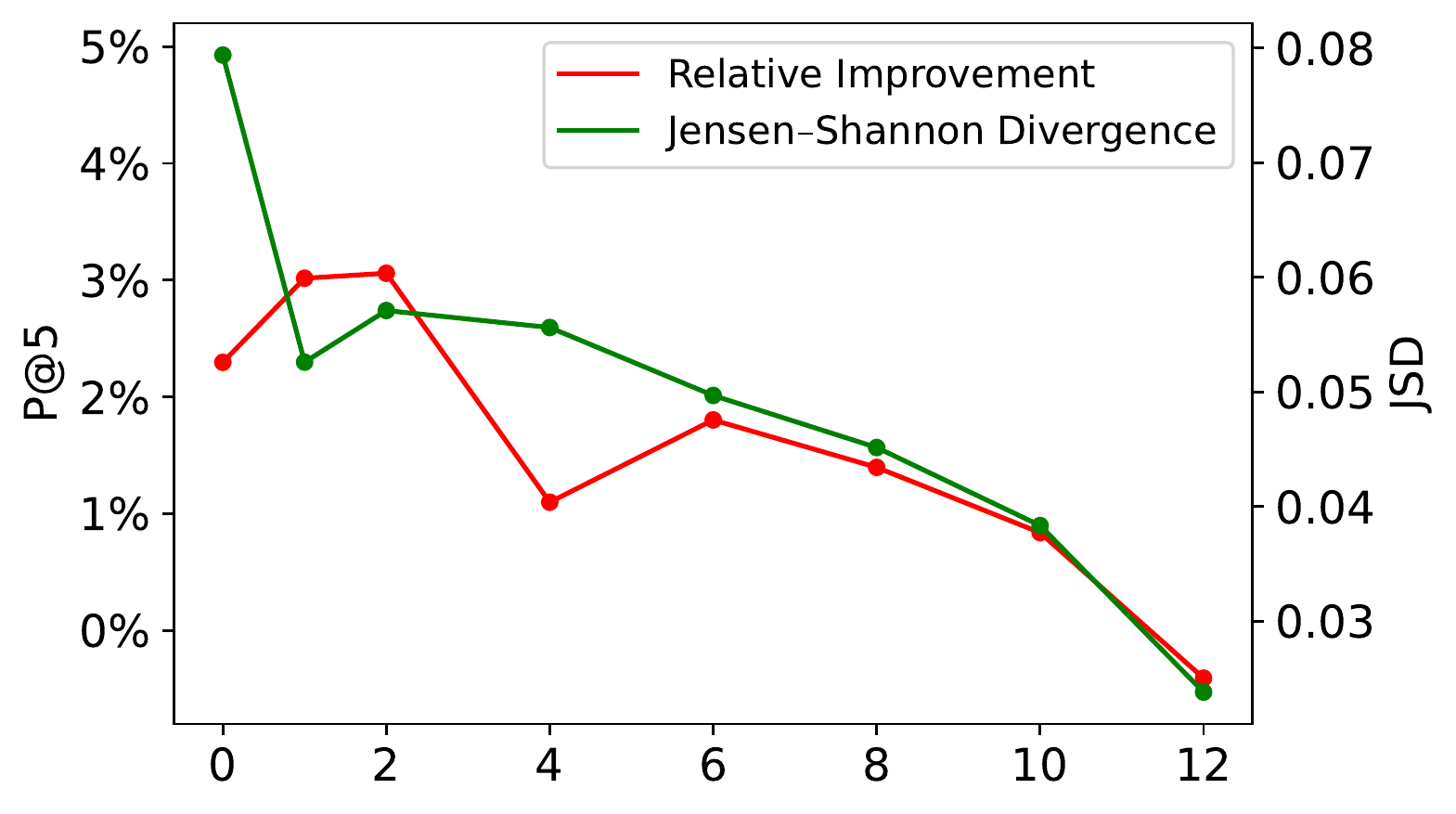}
  \caption{(EURLex-4K) the relative improvement of classification of \ourmodel over the global feature, and the JSD between predicted label distributions with global and local feature. We fix the global classifier and use the local feature from different layers.}
  \label{fig:ablation_jac}
\end{figure}
Our hypothesis is: even if the token embedding at higher layer preserves the token meaning \footnote{After all the MLM is optimized to predict the word identity at the final layer of pre-trained Transformer.}, it becomes more contextualized after multiple layers of self-attention.
Consequently, querying from more contextualized embeddings makes the label harder to pick up the salient keywords information. 
In \figref{fig:ablation_jac}, 
we study the correlation between the relative improvement (red curve) of \ourmodel over the global classifier and the JSD (green curve) of the predicted label distributions by the local and global classifier. It reveals: 1) the distributions by local and global classifiers are more similar when a higher layer of word embeddings are used, and 2) a higher distribution similarity is correlated with a lower improvement of \ourmodel. This shows that querying from more contextualized word embeddings may degrade to querying the global embedding (as does the global classifier) and result in less information gain.


\section{Conclusion}
In this paper, we propose \ourmodel, a classification system integrating both the global and local features from the pre-trained Transformers. The global classifier uses [CLS] embedding as the summarization of document, and the local classifier uses the label-word attention to directly select salient part of texts for classification. Our model combines the two to capture different granularity of document semantics, which achieves superior or comparable performances over SOTA methods on the benchmark datasets.


\end{document}